\documentclass{bmvc2k}

\usepackage[capitalize]{cleveref}
\usepackage{subcaption}
\usepackage{multirow}
\usepackage{pgfplots}
\usepackage{booktabs}
\usepackage[skip=2pt]{caption} 
\pgfplotsset{compat=1.16}

\title{\hspace{0.4cm}DeepliteRT: Computer Vision at the Edge}

\addauthor{\hspace{0.4cm}Saad Ashfaq}{saad@deeplite.ai}{1}
\addauthor{\hspace{0.4cm}Alexander Hoffman}{alexander.hoffman@deeplite.ai}{1}
\addauthor{\hspace{0.4cm}Saptarshi Mitra}{saptarshi@deeplite.ai}{1}
\addauthor{\hspace{0.4cm}Sudhakar Sah}{sudhakar@deeplite.ai}{1}
\addauthor{\hspace{0.4cm}MohammadHossein AskariHemmat}{mohammad@deeplite.ai}{1}
\addauthor{\hspace{0.4cm}Ehsan Saboori}{ehsan@deeplite.ai}{1}

\addinstitution{
 Deeplite Inc.\\
 Toronto, Canada\\
}

\runninghead{Ashfaq et al.}{DeepliteRT: Computer Vision at the Edge}

\begin{document}

\maketitle

\begin{abstract}
The proliferation of edge devices has unlocked unprecedented opportunities for deep learning model deployment in computer vision applications. However, these complex models require considerable power, memory and compute resources that are typically not available on edge platforms. Ultra low-bit quantization presents an attractive solution to this problem by scaling down the model weights and activations from 32-bit to less than 8-bit. We implement highly optimized ultra low-bit convolution operators for ARM-based targets that outperform existing methods by up to 4.34$\times$. Our operator is implemented within Deeplite Runtime (DeepliteRT), an end-to-end solution for the compilation, tuning, and inference of ultra low-bit models on ARM devices. Compiler passes in DeepliteRT automatically convert a fake-quantized model in full precision to a compact ultra low-bit representation, easing the process of quantized model deployment on commodity hardware. We analyze the performance of DeepliteRT on classification and detection models against optimized 32-bit floating-point, 8-bit integer, and 2-bit baselines, achieving significant speedups of up to 2.20$\times$, 2.33$\times$ and 2.17$\times$, respectively.

\end{abstract}

\section{Introduction}
\label{sec:intro}
Deep learning models for computer vision are being extensively deployed in various domains and industries due to substantial improvements in the accuracy of deep convolutional neural networks (CNNs). CNN architectures including VGG \cite{vgg}, ResNet \cite{resnet}, Inception \cite{inceptionv3}, DenseNet \cite{densenet} and YOLO \cite{yolov3} have demonstrated exceptional performance on image classification and object detection tasks. The widespread adoption of deep learning solutions in computer vision has also coincided with the growth of edge computing \cite{edge-computing}, promising the potential of bringing machine learning to low-power edge devices. However, the enhancements in CNN model accuracy have come at the expense of increased model complexity leading to high power, compute, memory, and storage requirements, making such models highly impractical for most use cases on resource-constrained edge devices.

Several compression techniques \cite{compression-survey} \cite{quant-survey} \cite{pruning-survey} have been explored to tackle this problem with the goal of decreasing model size while maintaining the baseline accuracy. Quantization is one such approach that realizes this goal by reducing the scale of model weights and activations from 32-bit floating-point (FP32) to lower precision representations. In addition to model compression, quantization also offers the benefits of fewer memory accesses, lower latency, and improved energy efficiency. 8-bit integer (INT8) has become the predominant bit-width for quantization and is widely supported in publicly available machine learning frameworks \cite{tensorflow2015-whitepaper} \cite{pytorch} that perform quantization-aware training (QAT) and in open-source inference engines \cite{onnxruntime} \cite{TFlite} that execute the quantized models on commodity hardware. Recent advances have also been made in ultra low-bit quantization where the model weights and activations are quantized to less than 8 bits of precision. Using methods such as LSQ \cite{LSQ}, a 2-bit quantized model can achieve a compression rate of up to 16$\times$ with an accuracy drop of less than a few percent relative to the FP32 baseline. Moreover, compute-intensive nodes in the network, including dense and convolution layers, can also utilize inexpensive bitwise operations to perform the dot products on extremely low-bit data. The significant compression and speedup resulting from ultra low-bit quantization make it a compelling choice for CNN deployment on edge devices. 

Deep learning workloads on CPU architectures in commodity off-the-shelf edge devices generally utilize Single Instruction, Multiple Data (SIMD) hardware units to perform operations on multiple inputs in parallel. INT8 inference can be easily performed as 8-bit SIMD instructions are available in the instruction set architectures (ISAs) of mainstream CPUs. On the other hand, ultra low-precision models necessitate operations on sub-8-bit data requiring custom kernel implementations since SIMD execution is generally unsupported on less than 8 bits. Moreover, the weights and activations are "fake-quantized" during the forward and backward passes of QAT. This means that the input values are rounded to a discrete set of floating-point values and all computations are still performed in full-precision during the training phase. In the case of INT8 quantization, model weights and activations in FP32 can be easily cast to standard 8-bit integer when exporting the quantized model for inference. However, for ultra low-bit quantization, the conversion to extremely low-precision can not be performed at this stage due to lack of support for sub-8-bit data types on the target platform. Typically, the machine learning framework used for training inserts custom operators for quantized layers such as convolution during model export after QAT. The inference engine then needs to parse these custom operators when loading the model, lower them to the corresponding ultra low-bit kernels based on the quantized layer, and pack the fake-quantized inputs in ultra low-bit data structures. These modifications required in both the training and the inference paths make it extremely challenging to deploy ultra low-bit models on real commodity hardware. 

To address these shortcomings in ultra low-bit pipelines, we introduce Deeplite Runtime (DeepliteRT), an end-to-end inference solution based on the TVM machine learning compiler stack \cite{tvm-endtoend}, that offers state-of-the-art performance and framework-agnostic deployment of ultra low-bit models on ARM CPUs. We implement an ultra low-bit convolution operator that improves upon the performance of the TVM bit-serial kernel \cite{tvm-lowprecision-cowan} \cite{tvm-quantkernels-cowan} by up to 4.34$\times$. We provide 32-bit ARMv7 and 64-bit ARMv8 bit-serial kernels making ultra low-bit CNN inference possible on globally pervasive ARM-based edge devices. We define compiler passes to automatically convert standard convolution layers into ultra low-bit operators and to efficiently pack full-precision data into compact ultra low-bit representations. These passes enable fake ultra low-bit quantized models trained with various ML frameworks to be executed on ARM CPUs without any additional changes in the training and inference paths. With support for mixed precision inference, layers in the network that are sensitive to quantization can be kept at higher precision (FP32, INT8, etc.) while insensitive layers can be reduced to ultra low-bit in order to minimize the accuracy drop resulting from quantizing all layers in the model. To summarize, this paper makes the following contributions:

\begin{itemize}
    \item We implement high performance bit-serial convolution kernels that achieve a speedup of up to 4.34$\times$ over existing ultra low-bit methods on ARM-based platforms.
    \item We present DeepliteRT, a compiler and runtime package for ultra low-bit inference on ARM CPUs. DeepliteRT automates the process of converting fake-quantized convolution layers from different machine learning frameworks used for quantization-aware training into ultra low-bit convolution kernels. Quantized models can be exported with the weights and activations still in full-precision without the need for custom operator definitions as compiler passes in DeepliteRT can handle the necessary casting, layout transforms and operator conversions during compilation. DeepliteRT provides a framework-agnostic end-to-end solution for ultra low-bit CNN deployment on edge devices eliminating the need to modify any code in the inference or runtime path.
    \item We perform a comprehensive evaluation of DeepliteRT on classification and detection models for both ARMv7 and ARMv8 targets, achieving significant performance improvements of up to 2.20$\times$, 2.33$\times$ and 2.17$\times$ over highly optimized FP32, INT8 and ultra low-bit baselines, respectively.
\end{itemize}





\begin{table}
    \scriptsize
    \centering
    \caption{2-bit accuracy on ImageNet with different QAT methods \cite{PACT} \cite{LQ-NETS} \cite{QIL} \cite{PACT-SAWB} \cite{LSQ}.}
    \label{accuracies}
    \begin{tabular}{lcccccc}
        \toprule
        \multirow{2}{*}{\textbf{Model}} & 
        \multirow{2}{*}{\textbf{Top-1}} &
        \multicolumn{5}{c}{\textbf{Top-1 Accuracy@2-bit}} \\ \cmidrule(lr){3-7}
        & \textbf{Accuracy@32-bit} & PACT (2018) & LQ-NET (2018) & QIL (2019) & PACT-SAWB (2019) & LSQ (2020) \\
        \midrule
        ResNet18 & 70.5\% & 64.4\% & 65.2\% & 65.7\% & 67.0\% & 67.9\%  \\
        ResNet50 & 76.9\% & 72.2\% & 71.5\% & & 74.2\% & 74.6\% \\ 
        \bottomrule
    \end{tabular}
\end{table}

\section{Related Work}
\label{sec:related}
\subsection{Ultra Low-bit Quantization}

Quantization methods can be broadly categorized into uniform and non-uniform as well as quantization-aware training (QAT) and post-training quantization (PTQ). Uniform quantization refers to the case where the floating-point weights are quantized to integer values with a linear scaling from the integer to floating-point domain. The benefit of these methods is that operations can be performed in the integer domain and quickly converted to the floating-point domain via multiplication of a scaling factor. Non-uniform quantization removes this restriction, allowing for more flexibility in the mapping from floating-point to integer data.

QAT quantizes weights and activations while training the model to better simulate the model's performance after quantized deployment. PTQ methods train a full-precision model without regard for quantization, and then quantize the model with minimal access to the training dataset. State-of-the-art ultra low-bit quantization methods, shown in \cref{accuracies}, make use of QAT to offset the loss of precision when reducing precision to less than 8 bits. LSQ \cite{LSQ} is a simple yet effective quantization method which takes advantage of both uniform quantization and QAT to quantize models to as low as 2 bits with minimal accuracy degradation. For example, ResNet18 quantized to 2 bits with LSQ only incurs a 2.4\% drop in accuracy relative to full-precision, but offers 16$\times$ compression per quantized layer.

\subsection{Ultra Low-bit Inference}
Most previous works on sub-8-bit inference on CPU architectures utilize the bit-serial method \cite{tvm-lowprecision-cowan} \cite{tvm-quantkernels-cowan} for dot product computation. Considering binary vectors with unipolar (unsigned) encoding where each input value is either 0 or 1, the bit-serial dot product is given by \cref{eqn:binarydotunipolar}. A bit-wise AND operation gives the element-wise product of the binary inputs and the popcount operation, that counts the number of bits set to 1, performs the accumulation. The binary case can easily be extended to larger bit-widths by slicing the inputs into binary vectors and performing a summation of the bit-serial dot products over all possible bit-sliced combinations. The corresponding equation for an M-bit weight and an N-bit activation vector is given in \cref{eqn:bitserialdotunipolar} where operations are performed across bit-planes ($w_{m}$ and $a_{n}$).
\begin{subequations}
    \begin{equation}
    \label{eqn:binarydotunipolar}
        \Vec{w} \cdot \Vec{a} = popcount(\Vec{w}\And\Vec{a})
    \end{equation}
    \begin{equation}
    \label{eqn:bitserialdotunipolar}
        \Vec{w} \cdot \Vec{a} = \sum_{m=0}^{M - 1}\sum_{n=0}^{N - 1}(popcount(\Vec{w_m}\And\Vec{a_n})) << (n + m)
    \end{equation}
\end{subequations}

This bit-serial approach is implemented within TVM for dense and convolution layers in \cite{tvm-lowprecision-cowan} and \cite{tvm-quantkernels-cowan} with an average speedup of 1.9$\times$ for a 2-bit ResNet18 network over an optimized FP32 baseline on the ARM Cortex-A53 CPU in the Raspberry Pi 3B. Riptide \cite{Riptide} also uses the bit-serial kernels in TVM along with fusion, vectorization and tiling optimizations for binary networks to achieve considerable latency improvements over full-precision models on the Cortex-A53. Bitflow \cite{bitflow} presents another bit-serial implementation of a binary VGG network for Intel CPUs that is even faster than the corresponding full-precision CNN tested on a high-performance GPU. There have also been initiatives in this space that are not based on the bit-serial method including ULPPACK \cite{ULPPACK}, BiQGEMM \cite{BiQGEMM} and DeepGEMM \cite{ganji2023deepgemm}.

\section{Bit-serial Convolution}


\begin{figure}
    \centering
    \includegraphics[width=\linewidth]{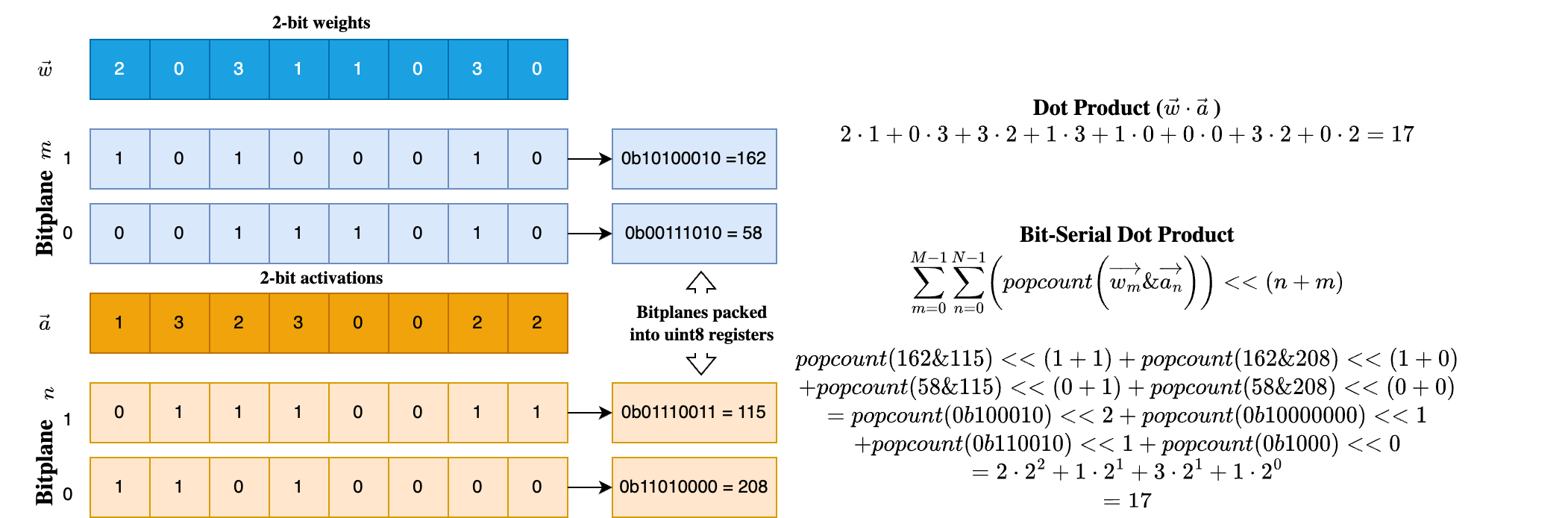}
    \caption{Input weight and activation values are sliced into bitplanes and bitpacked within unsigned 8-bit integers enabling dot product calculation using bitwise operations.}
    \label{fig:bitpacking}
\end{figure}
\label{sec:bitserial}
\subsection{Bitpacking}
Binary quantization approaches \cite{bnn-survey} can result in an unacceptable accuracy loss due to the use of a single bit for weight and activation values. To counter this, the bit-serial method can be extended to multiple bits by slicing the input weights and activation into separate bitplanes depending on the bit-width. This is illustrated in \cref{fig:bitpacking} for the 2A2W configuration (2 bits for activations and 2 bits for weights). Each value in the input data is first broken down into its constituent bits, creating bitplanes at every bit position. A bitplane holds the corresponding bit from different input values; for instance, bitplane 0 for weights stores the least significant bits across the weight values. Bitplanes can be compactly stored into standard data types such as 8-bit unsigned integers through the process of bitpacking. Assuming unipolar encoding for the 2-bit weights and activations, the bit-serial dot product can then be computed using \cref{eqn:bitserialdotunipolar} producing the same result as a standard dot product as shown in \cref{fig:bitpacking}. Based on our experiments, the bitpacking operation is not a major bottleneck consuming only 2-4\% of the overall execution time in the bit-serial computation.

\subsection{Optimized bit-serial dot product}
\label{subsec:dlrtbitserial}
\cref{eqn:bitserialdotunipolar} assumes a unipolar encoding scheme with unsigned values for both weights and activations. Recent works \cite{LSQ} \cite{PACT-SAWB} typically employ a hybrid unipolar-bipolar scheme with unipolar activations and bipolar (signed) weights producing quantized models with higher accuracy. The {\tt nn.bitserial\_conv2d} operator in TVM implements a convolution kernel for this hybrid scheme that calculates the bit-serial dot product as shown in \cref{eqn:bitserialdotbipolar}, providing an open-source SOTA baseline for comparison with our work. 

\begin{equation}
    \label{eqn:bitserialdotbipolar}
    \Vec{w} \cdot \Vec{a} = \sum_{m=0}^{M - 1}\sum_{n=0}^{N - 1}(popcount(\Vec{w_m}\And\Vec{a_n}) - popcount(\neg\Vec{w_m}\And\Vec{a_n})) << (n + m)
\end{equation}

Compared to the purely unipolar case in \cref{eqn:bitserialdotunipolar}, this version doubles the number of popcount instructions adding considerable latency to the dot product calculations. Moreover, the weights can not take on the value 0 since this bipolar scheme distributes the quantization levels around 0. For example, in the case of 2 bits, each weight value will lie in the discrete set \{-3, -1, 1, 3\}. Such a representation introduces error when quantizing zero values, which is particularly harmful for common operations such as zero-padding and ReLU \cite{quant-whitepaper}.

\begin{figure}
    \centering
    \begin{subfigure}{0.45\textwidth}
        \centering
        \captionsetup{justification=centering}
        \includegraphics[width=0.75\linewidth]{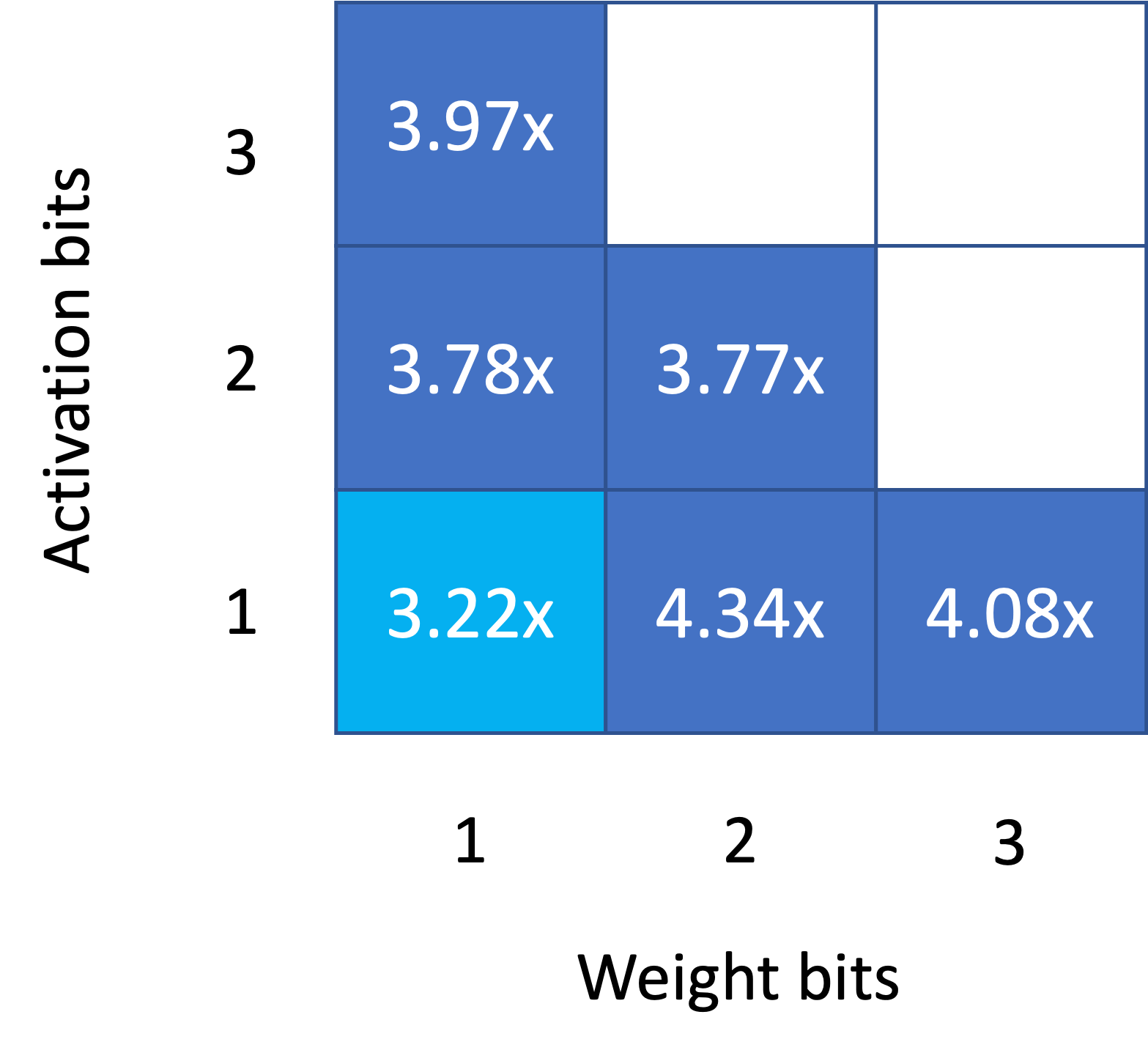}
        \caption{Speedup on second layer of ResNet18 across different bit-widths.}
        \label{fig:dlrt tvm op speedup}
    \end{subfigure}
    \begin{subfigure}{0.45\textwidth}
        \centering
        \captionsetup{justification=centering}
        \includegraphics[width=0.75\linewidth]{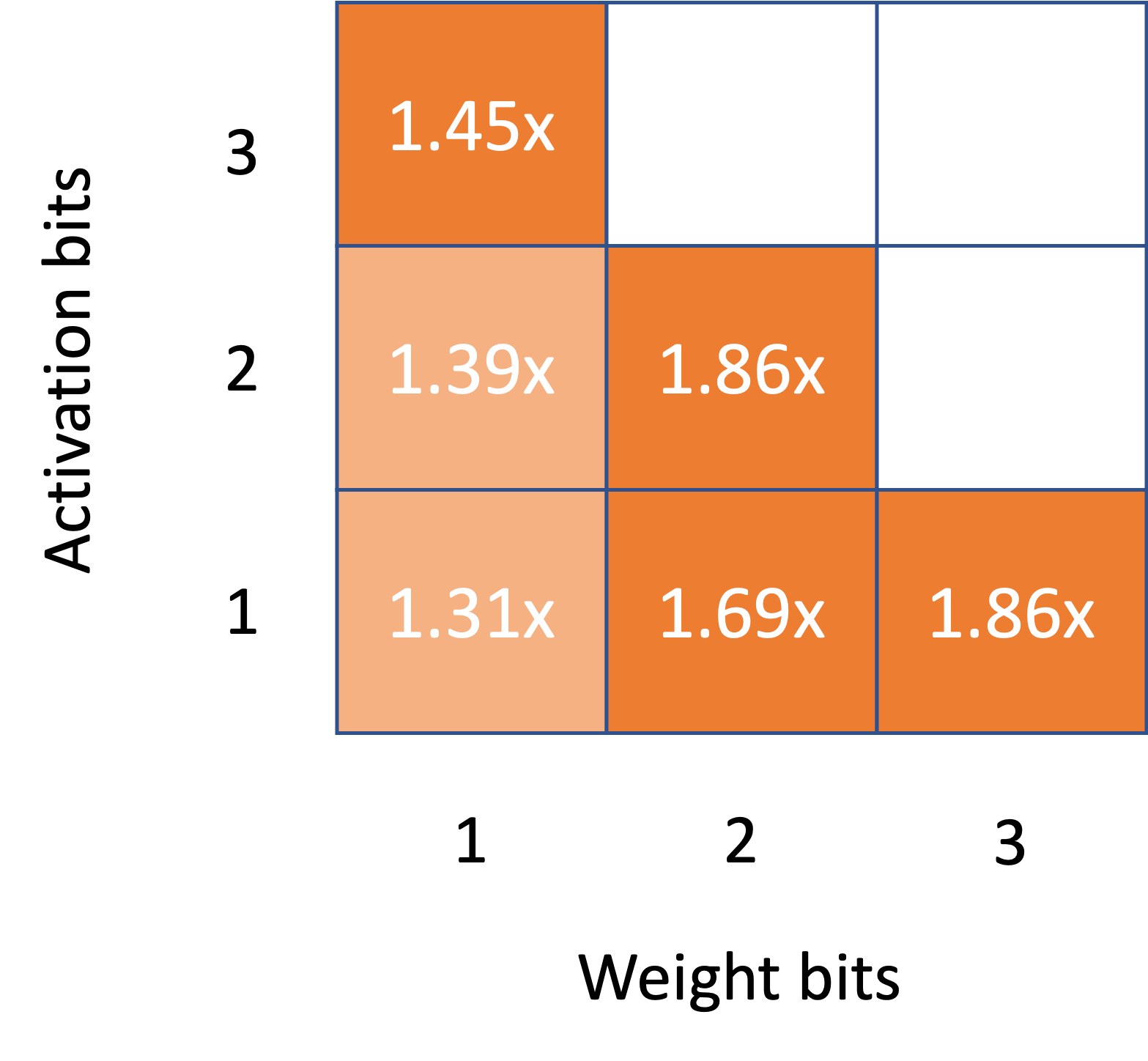}
        \caption{Speedup on ResNet18 model across different bit-widths.}
        \label{fig:dlrt tvm model speedup}
    \end{subfigure}
    \caption{Operator level and end-to-end speedups of {\tt dlrt\_bitserial\_conv2d} over TVM's {\tt nn.bitserial\_conv2d} on the Raspberry Pi 4B running in 32-bit mode.}
    \label{fig:dlrt tvm speedup}
\end{figure}

To address these drawbacks, we propose a novel bit-serial computation method in \cref{eqn:bitserialdotdlrt} for the hybrid scheme. Our approach reduces the number of popcount operations per dot product to one. It also requires the same number of overall instructions as the unipolar variant except for the most significant weight bit which has a slight overhead due to a constant multiplication. Our scheme also enables zero mapping of the signed weight values. For instance, 2-bit weights now fall in the set \{-2, -1, 0, 1\} providing compatibility with high accuracy quantization techniques such as LSQ that require zero mapping for the weights. This bit-serial dot product is the building block of our bit-serial convolution operator {\tt dlrt\_bitserial\_conv2d}. With optimizations in kernel and data vectorization, loop reordering, and parallelization, {\tt dlrt\_bitserial\_conv2d} achieves substantial performance uplifts over TVM's {\tt nn.bitserial\_conv2d} as shown in \cref{fig:dlrt tvm speedup}.
\begin{equation}
    \label{eqn:bitserialdotdlrt}
    \Vec{w} \cdot \Vec{a} = 
    \begin{cases}
    -1 \times \sum_{n=0}^{N - 1}(popcount(\Vec{w}_{M-1}\And\Vec{a_n})) << (n + m), & \text{if } m = M - 1\\
    \sum_{m=0}^{M - 1}\sum_{n=0}^{N - 1}(popcount(\Vec{w_m}\And\Vec{a_n})) << (n + m), & \text{otherwise}
    \end{cases}
\end{equation}

As opposed to the {\tt nn.bitserial\_conv2d} kernel that is only defined for ARMv7, we implement both 32-bit and 64-bit {\tt dlrt\_bitserial\_conv2d} kernels enabling deployment on a broader range of 32-bit ARMv7 and 64-bit ARMv8 platforms.

\section{DeepliteRT}
\label{sec:dlrt}
Machine learning frameworks used for ultra low-precision QAT such as PyTorch \cite{pytorch} and TensorFlow \cite{tensorflow2015-whitepaper} produce quantized models with extra operators relative to the full-precision network to handle the quantization and dequantization of model weights and activations. Assuming uniform quantization, these operators including addition, subtraction, division, multiplication, clipping and rounding are generally used to convert the floating-point data to integer before quantized layers and integer data back to floating-point after quantized layers. Inference engines such as ONNX Runtime \cite{onnxruntime} offer native support for these operators as they act on standard data types (FP32, INT16, INT8, etc.). However, quantized nodes such as convolution and dense layers are typically fake-quantized during QAT, restricting the weights and activations to a discrete set but still storing them in FP32. To realize ultra low-bit deployment on target hardware, custom operators and attributes for these layers have to be added by the ML framework which need to be then parsed and lowered to corresponding low-level kernels by the inference engine. These modifications in the ML and runtime frameworks require some level of expertise in both training and inference domains. Moreover, the changes made for one ML framework are not portable to a different framework, making quantized ultra low-bit model deployment inaccessible to most practitioners.

\begin{figure}
    \centering
    \includegraphics[width=0.95\linewidth]{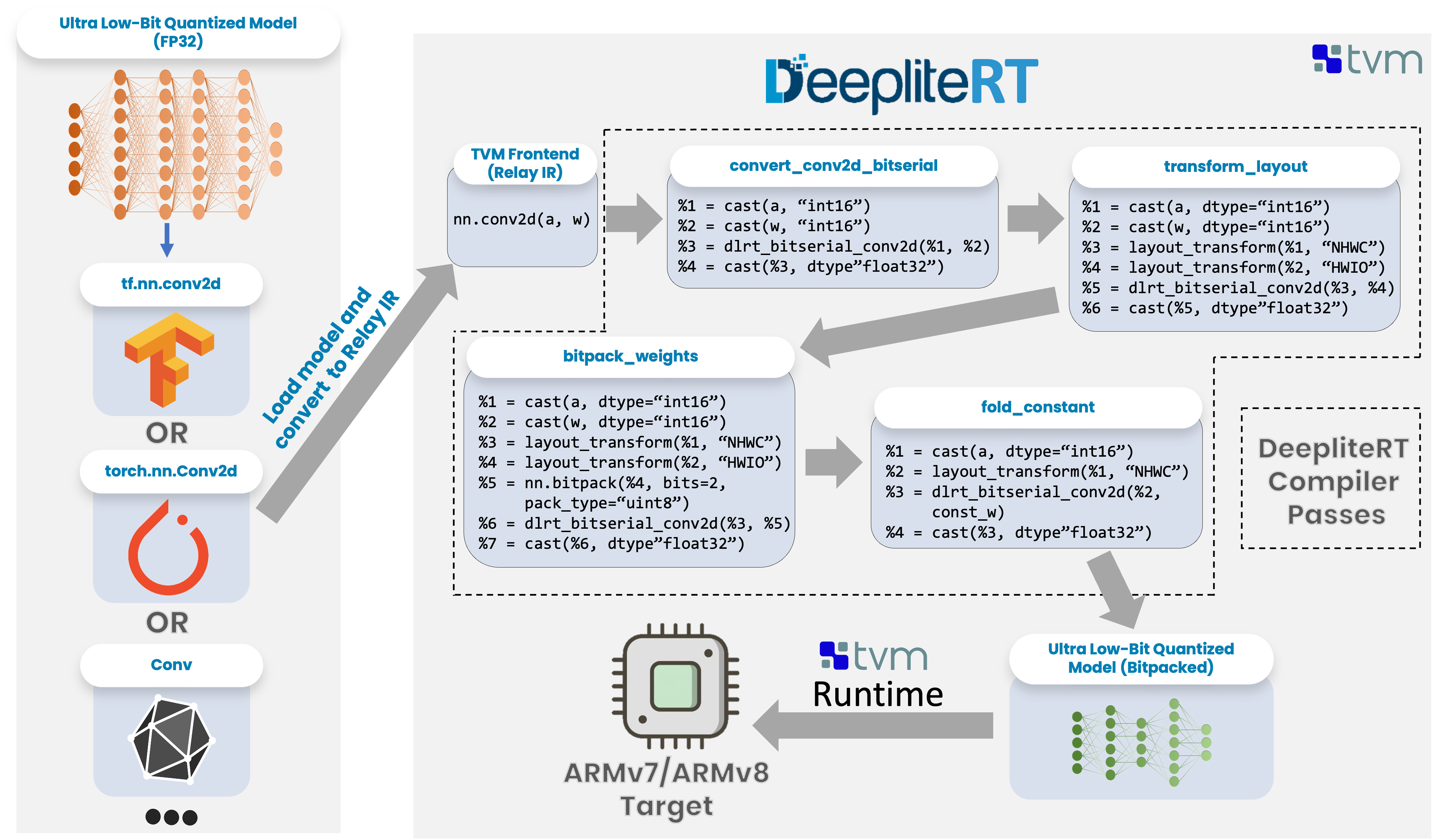}
    \caption{DeepliteRT converts fake-quantized convolution layers from models in different formats to optimized ultra low-bit convolution operators through a series of compiler passes. The passes replace {\tt nn.conv2d} with {\tt dlrt\_bitserial\_conv2d}, bitpack the weights in ultra low-bit, and cast and transform the layouts of data as required. The resulting compiled model can be deployed on ARMv7 and ARMv8 CPUs via TVM runtime. }
    \label{fig:deeplitert}
\end{figure}

DeepliteRT is an inference solution that defines custom compiler passes in the TVM machine learning compiler to transform fake-quantized models into compact ultra low-precision networks. ML practitioners can perform QAT in any framework of choice and simply compile the resulting fake-quantized model with DeepliteRT for easy deployment on ARM-based targets with TVM runtime.  DeepliteRT includes our optimized bit-serial convolution operator detailed in \cref{subsec:dlrtbitserial}. It also supports mixed precison deployment allowing quantization-sensitive layers to be kept at higher precision and insensitive layers at ultra low-precision. With high-level APIs in both Python and C++, DeepliteRT can be easily integrated in applications on edge devices for quantized model compilation, tuning and inference.

\subsection{Compiler passes}
{\tt nn.conv2d} is the operator for 2D convolution in TVM's Relay IR. Convolution layers from models trained with different ML frameworks are internally converted into {\tt nn.conv2d} by the appropriate frontend. For instance, {\tt tf.nn.conv2d} from a TensorFlow model,\\ {\tt torch.nn.Conv2d} from a PyTorch model and {\tt Conv} from an ONNX model are all translated to {\tt nn.conv2d}. We define a sequence of compiler passes in DeepliteRT to convert a fake quantized convolution layer represented by {\tt nn.conv2d} in Relay IR into our optimized bit-serial convolution operator {\tt dlrt\_bitserial\_conv2d} as shown in \cref{fig:deeplitert}.\\
\textbf{convert\_conv2d\_bitserial:} This custom pass converts {\tt nn.conv2d} nodes for quantized layers into {\tt dlrt\_bitserial\_conv2d} nodes in the IR. It also casts the input weights and activations into integer and the resulting convolution output back to floating-point.\\
\textbf{transform\_layout:} This pass is invoked to change the layout for activations to NHWC and the layout for weights to HWIO as required by the low-level \\{\tt dlrt\_bitserial\_conv2d} kernel. The transformation is only performed if the activations and/or weights are not already in the required layouts.\\
\textbf{bitpack\_weights:} This custom pass adds {\tt nn.bitpack} operators in the Relay IR for the bitpacking of weights during compilation prior to bit-serial convolution. The bitpacking of activations is handled by the {\tt dlrt\_bitserial\_conv2d} operator during inference since the activation values are not available offline.\\
\textbf{fold\_constant:} This pass is used to perform all the computations on weights during compilation as they are compile-time constants. The result of casting the weights to integer, transforming their layout and bitpacking them is then simply passed as a constant to the {\tt dlrt\_bitserial\_conv2d} operator.

\begin{table}
    \scriptsize
    \centering
    \caption{End-to-end latencies (ms) and speedups of DeepliteRT 2A2W over TVM FP32, ONNX Runtime INT8 and TVM bit-serial 2A2W baselines.}
    \label{latencies}
    \begin{tabular}{lcccccccc}
        \toprule
        \multirow{2}{*}{\textbf{Model}} & 
        \multicolumn{4}{c}{\textbf{Raspberry Pi 4B - 32-bit ARMv7}} & 
        \multicolumn{4}{c}{\textbf{Raspberry Pi 4B - 64-bit ARMv8}}\\  \cmidrule(lr){2-5} \cmidrule(lr){6-9}
        & FP32 & INT8 & 2A2W & 2A2W (Ours) & FP32 & INT8 & 2A2W & 2A2W (Ours) \\
        \midrule
        ResNet18 & 149.29 & 145.44 & 130.92 & 70.32 & 110.94 & 91.13 & 123.28 & 67.13  \\ 
        ResNet50 & 433.19 & 326.49 & 311.8 & 196.79 & 315.03 & 203.56 & 295.96 & 197.91  \\ 
        ResNet101 & - & 558.47 & 487.96 & 325.37 & 545.01 & 378.27 & 471.71 & 319.09 \\
        VGG19 & - & 1399 & 1003 & 654.69 & - & 922.28 & 962.65 & 636.79 \\ 
        InceptionV3 & 312.82 & 245.16 & 357.77 & 165.05 & 218.18 & 151.55 & 340.82 & 164.62 \\
        DenseNet121 & 387.98 & 589.03 & 296.27 & 252.65 & 302.50 & 261.94 & 269.91 & 227.05 \\
        \midrule
        VGG16-SSD300 & 1671 & 2310 & 1780 & 1190 & 1547 & 1462 & 1631 & 1060 \\
        YOLOv5s & 219.72 & 197.27 & 135.64 & 100.32 & 169.93 & 113.5 & 130.03 & 97.49 \\
        \midrule
        Average speedup & 1.89$\times$ & 1.91$\times$ & 1.58$\times$ & - & 1.54$\times$ & 1.20$\times$ & 1.56$\times$ & -\\
        Minimum speedup & 1.40$\times$ & 1.49$\times$ & 1.17$\times$ & - & 1.32$\times$ & 0.92$\times$ & 1.19$\times$ & -\\
        Maximum speedup & 2.20$\times$ & 2.33$\times$ & 2.17$\times$ & - & 1.71$\times$ & 1.45$\times$ & 2.07$\times$ & -\\
        \bottomrule
    \end{tabular}
\end{table}

\subsection{Mixed precision support}
In the default case, DeepliteRT converts all convolution layers except the first to bit-serial operators using the specified bit-width. However, quantizing all the layers to ultra low-bit can result in severe accuracy degradation. This can be countered with mixed precision quantization by choosing different precisions across layers using methods such as HAWQ-V3 \cite{hawq-v3} for accuracy preservation. DeepliteRT provides mixed precision inference by accepting a configuration file as input that specifies the quantization parameters per layer including activation bit-width, weight bit-width and encoding scheme. This per-layer information is passed to the \textbf{convert\_conv2d\_bitserial} pass to selectively offload convolution layers to ultra low-bit with the provided bit-widths and keep other layers in full-precision as required.

\section{Evaluation}
\label{sec:results}
We evaluate classification and detection models on a Rasbperry Pi 4B (4$\times$ARM Cortex-A72@1.8GHZ) device with 32-bit and 64-bit operating systems to enable ARMv7 and ARMv8 execution. We select TVM FP32 for the full-precision baseline as it significantly outperformed FP32 kernels in ONNX Runtime and TensorFlow Lite \cite{TFlite} in our experiments. TVM does not offer an optimized INT8 operator so we choose ONNX Runtime for INT8 experiments due to its high performance 8-bit kernels. Finally, we use the TVM 2A2W configuration based on the {\tt nn.bitserial\_conv2d} operator for ultra low-bit experiments; we also port this operator to ARMv8 to establish the 64-bit 2A2W baseline. All models deployed with TVM and DeepliteRT were tuned using AutoTVM \cite{tvm-learningtoopt} with 1500 trials.

\begin{figure}[t]
    \centering
    \begin{subfigure}{0.45\textwidth}
        \centering
        \includegraphics[width=\linewidth]{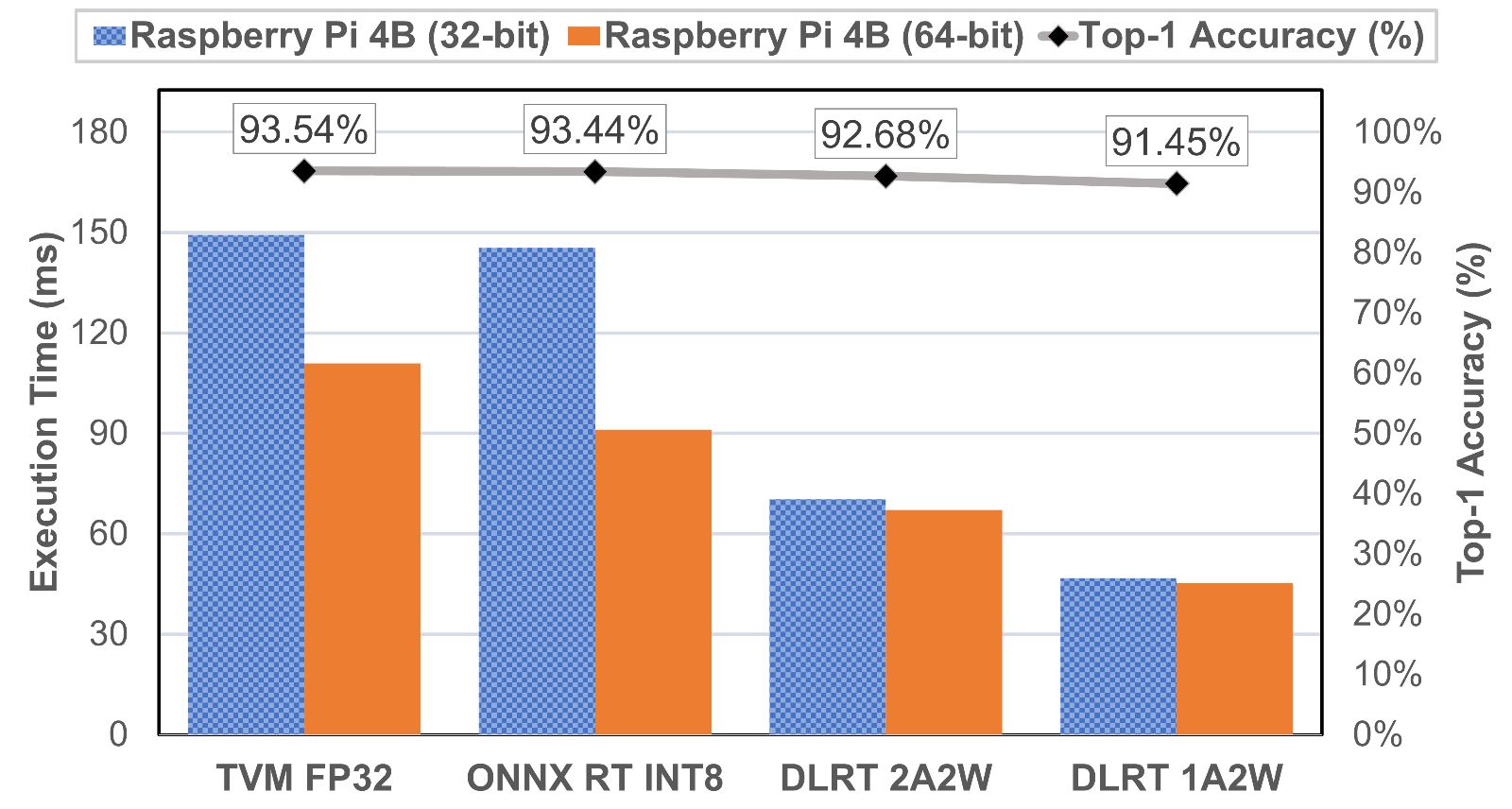}
        \caption{ResNet18 on VWW.}
        \label{fig:resnet 18 vww}
    \end{subfigure}
    \begin{subfigure}{0.45\textwidth}
        \centering
        \includegraphics[width=\linewidth]{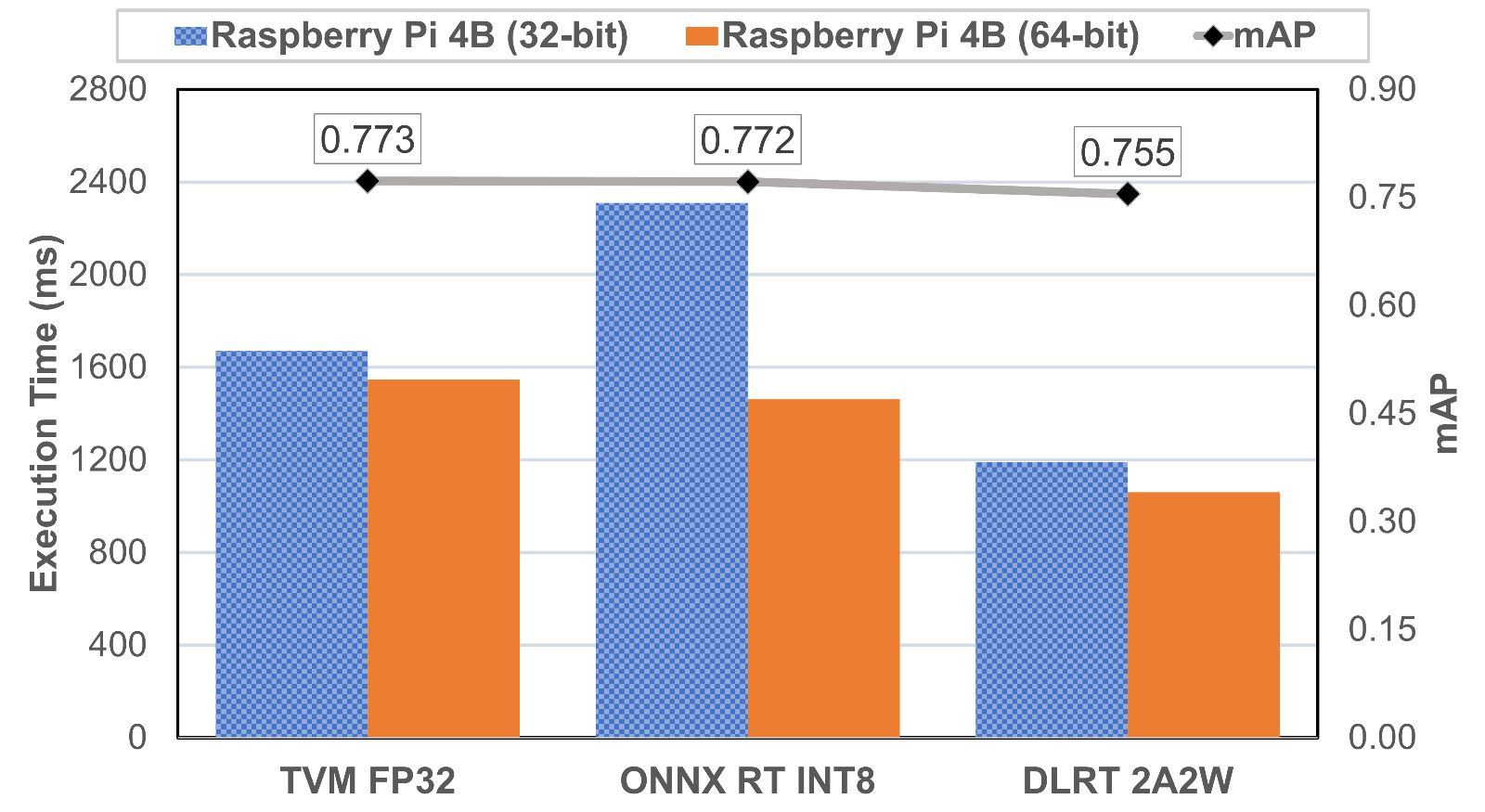}
        \caption{VGG16-SSD300 on VOC.}
        \label{fig:vgg16 ssd300 voc}
    \end{subfigure}
    \caption{Trade off between ultra low-bit model accuracy and performance.}
    \label{fig:accuracy performance tradeoff}
\end{figure}

\subsection{End-to-end performance}
\cref{latencies} reports the end-to-end latencies and speedups for  classification and detection models. The average, minimum and maximum numbers represent the speedups realized with DeepliteRT over the TVM FP32, ONNX Runtime INT8 or TVM 2A2W results in the same column. Some results for ResNet101 and VGG19 at FP32 are missing in the table as the device runs out of memory when loading full-precision model parameters. Interestingly, even though the TVM 2A2W configuration offers similar level of performance in 32-bit and 64-bit modes, it does not remain competitive in the latter case due to substantial performance uplifts for the FP32 and INT8 baselines with the ARMv8 ISA. In contrast, DeepliteRT offers leading performance for both ARMv7 and ARMv8 targets. On average, DeepliteRT realizes speedups of 1.89$\times$, 1.91$\times$ and 1.58$\times$ in 32-bit mode and 1.54$\times$, 1.20$\times$ and 1.56$\times$ in 64-bit mode over TVM FP32, ONNX Runtime INT8 and TVM 2A2W, respectively.

\begin{table}[h]
    \scriptsize
    \centering
    \caption{DeepliteRT latency (ms) on ResNet50 with mixed precision configurations.}
    \label{mixed precision}
    \begin{tabular}{lcccc} 
        \toprule
        \textbf{52 FP32} & \textbf{26 FP32 + 26 2A2W} & \textbf{52 2A2W} & \textbf{26 2A2W + 26 1A2W} & \textbf{52 1A2W}\\
        \midrule
        433.19 & 314.69 & 196.79 & 180.37 & 134.26 \\
        \bottomrule
    \end{tabular}
\end{table}

\subsection{Model accuracy and mixed precision}
SOTA for ultra low-bit quantization has progressed at a rapid pace as shown in \cref{accuracies}. We study the accuracy-performance tradeoff of ultra low-bit quantization using LSQ for a classification and detection model in \cref{fig:accuracy performance tradeoff}. ResNet18 trained on the VWW dataset \cite{vww_dataset} only incurs accuracy drops of 0.86\% and 2.09\% relative to the FP32 baseline with performance uplifts of up to 2.12$\times$ and 3.19$\times$ at 2A2W and 1A2W, respectively. Similarly, VGG16-SSD300 \cite{ssd300} trained on the VOC dataset \cite{voc_dataset} only sees a 0.18 loss in mAP at 2A2W while realizing a speedup of up to 1.46$\times$. The minor accuracy dips, substantial latency improvements and huge savings in model size make ultra low-bit networks an ideal fit for edge deployment. Moreover, mixed precision inference with DeepliteRT enables practitioners to easily explore this tradeoff between  accuracy and performance, as illustrated in \cref{mixed precision} for ResNet50, by varying the number of layers in FP32, 2A2W and 1A2W. An appropriate quantization configuration can be chosen based on model accuracy and latency measurements from the target.

\section{Conclusion}
\label{sec:conclusion}
We present an end-to-end inference solution in DeepliteRT for ML framework-agnostic deployment of ultra low-bit quantized models on 32-bit ARMv7 and 64-bit ARMv8 platforms. It implements compiler passes for the automatic conversion of fake-quantized networks in full-precision to compact representations in ultra low-bit, eliminating the need for custom modifications in the training and runtime components to enable inference at ultra low-precision. Using high-performance bit-serial convolution kernels, DeepliteRT outperforms highly optimized floating-point, integer, and ultra low-bit baselines on image classification and object detection models by up to 2.20$\times$, 2.33$\times$ and 2.17$\times$, respectively.

\bibliography{egbib}
\end{document}